\documentclass[10pt,twocolumn,letterpaper]{article}

\usepackage{cvpr}
\usepackage{times}
\usepackage{epsfig}
\usepackage{graphicx}
\usepackage{amsmath}
\usepackage{amssymb}

\usepackage{paralist}
\usepackage{subfigure}
\usepackage{multirow}
\usepackage{tikz}
\usepackage{diagbox}
\usepackage{enumitem}
\usepackage{multicol}
\usepackage{booktabs}

\usepackage{capt-of}

\usepackage[breaklinks=true,colorlinks,bookmarks=false]{hyperref}
\usetikzlibrary{shapes,arrows,shadows}

\definecolor{orange}{rgb}{1, 0.5, 0}
\definecolor{yellowGreen}{rgb}{0.75, 1, 0}
\definecolor{yellowGreen}{rgb}{0.75, 1, 0}
\definecolor{GreenBlue}{rgb}{0, 1, 0.5}

\tikzstyle{decision} = [diamond, draw, fill=blue!20, text width=4.5em, text badly centered, node distance=3cm, inner sep=0pt]
\tikzstyle{block} = [rectangle, draw, fill=blue!20, text width=5em, text centered, rounded corners, minimum height=4em]
\tikzstyle{line} = [draw, -latex']
\tikzstyle{cloud} = [draw, ellipse,fill=red!20, node distance=3cm, minimum height=2em]
\newcommand{\Aindent}{\leavevmode\hphantom{\textbf{A: }}}
\newcommand{\aindent}{\leavevmode\hphantom{(a) }}

\newcommand{\smallsim}{\smallsym{\mathrel}{\sim}}

\makeatletter
\newcommand{\smallsym}[2]{#1{\mathpalette\make@small@sym{#2}}}
\newcommand{\make@small@sym}[2]{%
  \vcenter{\hbox{$\m@th\downgrade@style#1#2$}}%
}
\newcommand{\downgrade@style}[1]{%
  \ifx#1\displaystyle\scriptstyle\else
    \ifx#1\textstyle\scriptstyle\else
      \scriptscriptstyle
  \fi\fi
}
\makeatother

\cvprfinalcopy 

\def\cvprPaperID{1847} 

\ifcvprfinal\fi
\begin{document}

\title{VQA with No Questions-Answers Training}

\author{Ben-Zion Vatashsky \hskip 4em  Shimon Ullman\\
Weizmann Institute of Science, Israel\\
{\tt\small vatashsky@gmail.com, shimon.ullman@weizmann.ac.il}
}

\maketitle
\thispagestyle{empty}

\vspace{-2ex}
\begin{abstract}

Methods for teaching machines to answer visual questions have made significant progress in recent years, but current methods still lack important human capabilities, including integrating new visual classes and concepts in a modular manner, providing explanations for the answers and handling new domains without explicit examples. We propose a novel method that consists of two main parts: generating a question graph representation, and an answering procedure, guided by the abstract structure of the question graph to invoke an extendable set of visual estimators. Training is performed for the language part and the visual part on their own, but unlike existing schemes, the method does not require any training using images with associated questions and answers. This approach is able to handle novel domains (extended question types and new object classes, properties and relations) as long as corresponding visual estimators are available. In addition, it can provide explanations to its answers and suggest alternatives when questions are not grounded in the image. We demonstrate that this approach achieves both high performance and domain extensibility without any questions-answers training.

\end{abstract}
\vspace{-2ex}

\begin{figure} [t]
\begin{centering}
\resizebox{\columnwidth}{!}{%
\begin{tabular}{llll}
\multicolumn{1}{c}{\subfigure{
  \includegraphics[totalheight=0.16\textheight]{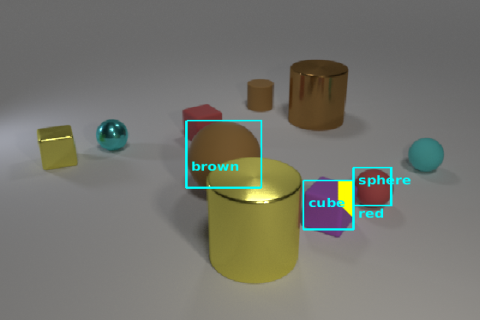} 
}} &
\multicolumn{1}{c}{\subfigure{
  \includegraphics[totalheight=0.16\textheight]{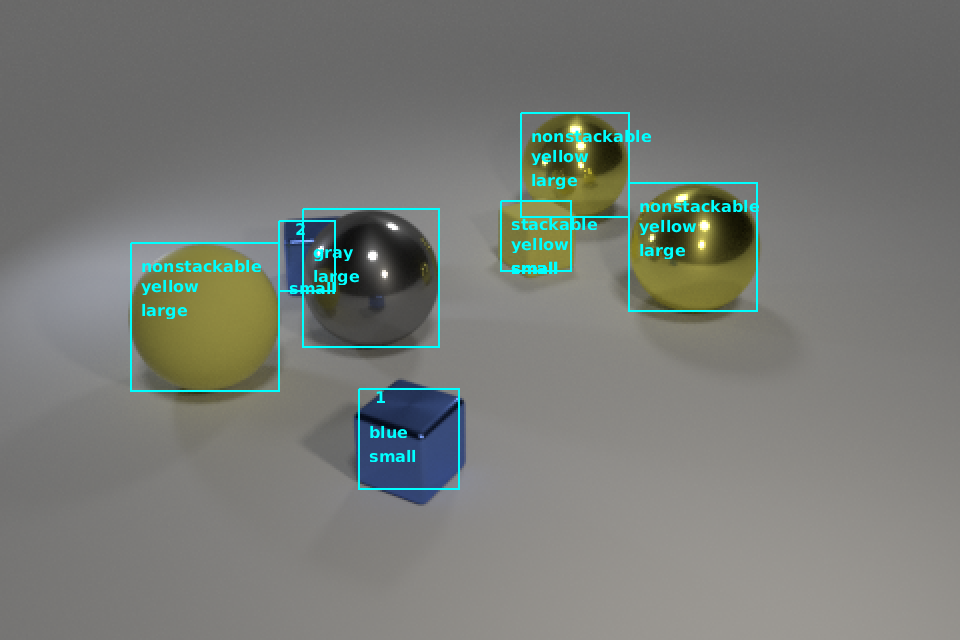} 
}}  \\
\textbf{Q}: What shape is the object  \textit{\color{red}{\underline{\color{black}closest to}}}  & \textbf{Q}: How many other objects are the same     \\  
\Aindent the red sphere?                                                                         & \Aindent    size as the yellow \textit{\color{red}{\underline{\color{black}stackable}}} object?  \\ 
\textbf{A}: cube                                                                                 & \textbf{A}:  2                                                 \\
\multicolumn{1}{c}{\subfigure{
  \includegraphics[totalheight=0.16\textheight]{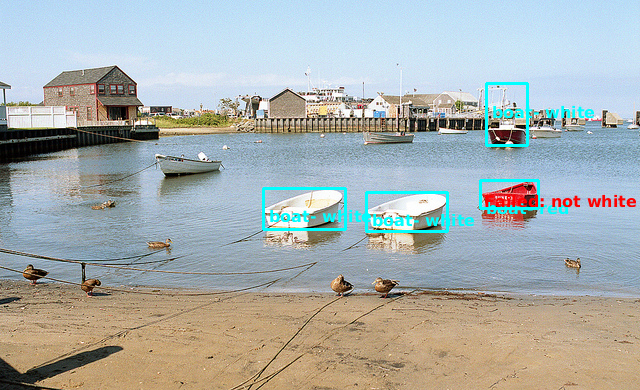} 
}} &
\multicolumn{1}{c}{\subfigure{
  \includegraphics[totalheight=0.16\textheight]{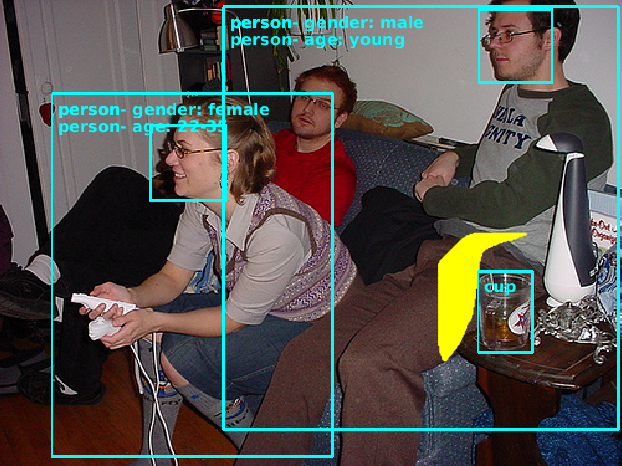} 
}}  \\
\textbf{Q}: Are all the boats white?                           & \textbf{Q}: \small{There is a person that is of a different} \\  
\textbf{A}: no \small{[full: There are not enough white boats} & \Aindent \small{gender than the young person}                \\ 
\Aindent \small{(failed due to a red boat)]}                   & \Aindent \small{closest to the cup; how old is he?}          \\
                                                               & \textbf{A}: 22-35

\end{tabular}
}
\caption[CLEVR new QA examples]{UnCoRd generalizes without QA training to novel properties and relations (top), and to real-world domain (bottom).}
\label{fig:clevr_new}
\end{centering}
\vspace{-3ex}
\end{figure}

\vspace{-1ex}
\begin{figure*}[t]
\begin{center}
\resizebox{2\columnwidth}{!}{%
\includegraphics{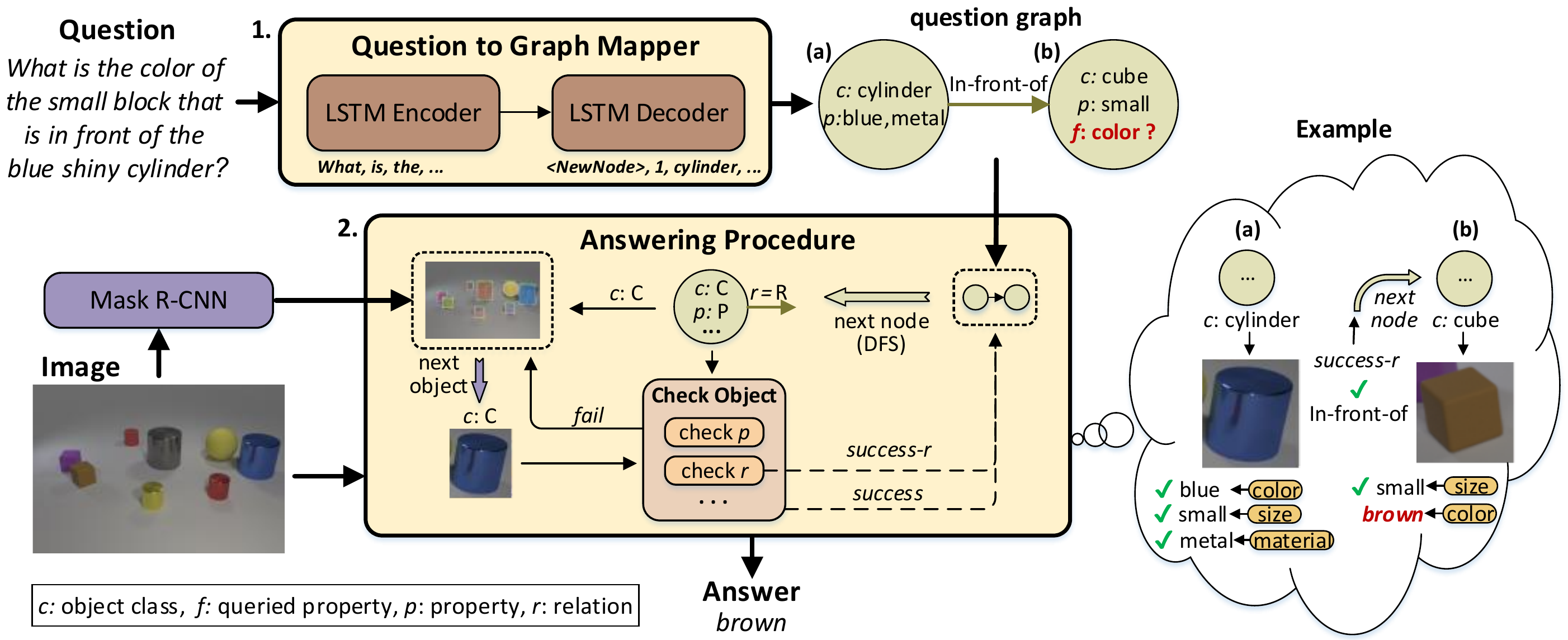}
}
\caption[scheme]{A schematic illustration of our method. The first stage (1) maps the question into a graph representation using a sequence-to-sequence \scalebox{0.94}[1.0]{LSTM} based model. At the second stage (2), the recursive answering procedure follows the graph, searching for a valid  assignment in the image. At each step, the handled node is set and objects (extracted using mask R-CNN) are examined according to the node's requirements (utilizing corresponding visual estimators). If succeeded, a new node is set (according to a DFS traversal) and the function is called again to handle the unassigned subgraph. The Example illustrates the flow: \scalebox{0.91}[1.0]{\textit{'check node (a)'} $\rightarrow$ \textit{'relation success'} $\rightarrow$ \textit{'check node (b)'} $\rightarrow$ \textit{answer}.}}
\label{fig:scheme}
\end{center}
\vspace{-4.5ex}
\end{figure*}

\section{Introduction}
\label{Introduction}
\vspace{-1ex}

Visual question answering is inspired by the remarkable human ability to answer specific questions on images, which may require analysis of subtle cues, along with the integration of prior knowledge and experience. The learning of new visual classes, properties and relations, can be easily integrated into the question-answering process.
Humans can elaborate on the answers they give, explain how they were derived, and why they failed to produce an adequate answer. Current approaches to handle VQA by a  machine  \cite{teney2017tips,srivastava2019visual,pandhre2017survey,wu2016visual,Kafle2016vqasurvey} take a different path, where most answering systems are trained directly to select an answer from common answers of a training set, based on fused image features (mostly using a pre-trained CNN \cite{hochreiter1997long}) and question features (mostly using an RNN).



The answering approach we take below is the first, as far as we know, that does not rely on any explicit question-answering training. It uses a process composed according to the question's structure, and applies a sequence of `visual estimators' for object detection and identifying a set of visual properties and relations.
Answering by our 'Understand, Compose and Respond' (UnCoRd) approach is divided into two stages (illustrated in Figure \ref{fig:scheme}). First, a graph representation is generated for the question, in terms of classes, properties and relations, supplemented with  quantifiers and logical connectives. An answering procedure then follows the question graph, and seeks either a single or multiple assignments of the classes, properties and relations in the graph to the image (Section \ref{sec:procedure}). The method is modular, extensible and uses intermediate results to provide elaborated answers, including alternatives to answers not grounded in the image, and notifying about unsupported categories. With an ability to handle extended domains, the UnCoRd approach demonstrates the potential to build a general answering scheme, not coupled to a specific dataset. 

Our work includes several novel contributions. First, a method that produces state-of-the-art results on the CLEVR dataset \cite{johnson2016clevr} without any questions-answers training. Second, we developed sequence-to-sequence based method, independent of images, to map questions into their graph representation. Third, we describe a formalism to represent a broad range of possible questions, with an algorithm that finds valid assignments of a question graph in the image, and provides an answer. Fourth, we present a model that can both perform well on CLEVR, as well as generalize to novel domains by just adding visual estimators (for objects, properties and relations) but without QA examples. Some examples are shown in Figure \ref{fig:clevr_new} (elaborated later in text).

\vspace{-1ex}
\section{Related Work}
\label{relatedWork}
\vspace{-1ex}

Current answering schemes are dominated by end-to-end methods, trained as multi-class classifiers. Many recent works focused on improving 
the image-question fused features \cite{akira16,ben2017mutan,yu2017multi,ben2019block,aioz_cti_iccv19}, attention mechanisms for selecting important features \cite{Yang2016CVPR,nguyen2018improved,lu2018co-attending,anderson2018bottom,Cadene_2019_CVPR,selvaraju2019taking,malinowski2018learning,hudson2018compositional}, including self and guided attention \cite{yu2019deep,guo2019graph,yu2019multimodal}, applying pre-trained networks \cite{tan2019lxmert,lu2019vilbert,zhou2019unified}, and incorporating outputs of other visual tasks \cite{Gupta_2017_ICCV,agrawal2018dont,desta2018,Teney_2017_CVPR,li2019relation,wu:acl19,kim:acl19,hudson2019learning}.
Some provide reasoning using "facts" extraction (\eg scene type) \cite{wang2017vqa}, image caption \cite{li2018vqa,aditya2018explicit,li2018tell} or by linking visual "facts" with question's logical form \cite{malinowski14nips,KrishnamurthyK13}. Other integrated external prior knowledge, by generating a query to a knowledge database \cite{wang2016fvqa,cao2019explainable}, fusing it in the representation \cite{Wu2016CVPR,li2017incorporating}, using a textual image description \cite{li2019visual} or by added loss terms \cite{teney2019incorporating}.
The language prior was addressed as well \cite{guo2019quantifying,grand2019adversarial,cadene2019rubi,wu2019self,ray2019convqa}.



Some methods use dynamic networks with architecture affected by the question \cite{gao2018question,perez2018film}. The Neural Module Networks (NMN) are dynamically composed out of sub-modules. Originally modules arrangement was based on the dependency parsing of the question \cite{andreas2016neural,andreas16naacl}, while later versions used supervised answering program learning \cite{johnson2017inferring,hu2017learning,mascharka2018transparency,suarez2018ddrprog}, including a probabilistic model \cite{vedantam2019probabilistic}. Note that the modules are trained only as components of an answering network for a specific dataset and do no function as independent visual estimators. One method \cite{nsvqa} performs full scene analysis in order to carry out the program. This method uses questions-answers training to learn the programs, hence cannot be extended by a simple addition of visual estimators. Moreover, performing full scene analysis (detecting all objects, properties and relations in the scene) may become infeasible for data less restricted than CLEVR (especially for relations). In our method, the answering process is guided by the question and does not perform a full scene analysis. It allows a flexible integration of additional visual capabilities (\eg novel object classes), providing elaborated answers and proposing alternatives.
These capacities are obtained without requiring any QA examples.
Current methods fit models to particular datasets and exploit inherent biases, which can lead to ignoring parts of the question/image, and to failures on novel domains and rephrasing \cite{agrawal2016analyzing,shah2019cycle}. In contrast to the modular approach we pursue, any adaptation or upgrade requires a full retraining. 

\vspace{-1ex}
\section{Method}
\label{sec:system}
\vspace{-1ex}

\subsection{Overview}
\label{sec:overview}
\vspace{-1ex}

In the formalism we use, a simple question without quantifiers can be transformed to an assertion about the image that may have free variables (\eg 'color' in 'what is the color of...'). The question is answered by finding an assignment to the image that will make the statement true, and retrieving the free variables. The quantifiers derived from the question require multiple true assignments (such as '5', 'all', etc.). The procedure we use seeks the required assignments and returns the desired answer. 
The answering process consists of two stages (see Figure \ref{fig:scheme} for a scheme):

\pgfdeclarelayer{background}
\pgfdeclarelayer{foreground}
\pgfsetlayers{background,main,foreground}


\tikzstyle{bp}  = [draw , fill=  green!20, text width=  5em, text centered, minimum height=2.5em, drop shadow]
\tikzstyle{vis} = [bp   , fill=   blue!20, text width=3.6em, minimum height=1.4em]
\tikzstyle{ann} = [above, text width= 5em, text centered]
\tikzstyle{op}  = [bp   , fill=    red!20, text width= 8em, minimum height=11em, rounded corners]
\tikzstyle{ex}  = [op   , fill= orange!20, text width= 5em, minimum height=5em]
\tikzstyle{im}  = [ex   , minimum height=6.7em, text width= 5.6em]
\tikzstyle{ch}  = [vis  , fill=  brown!15]
\tikzstyle{ec}  = [ch   , fill=  brown!25]
\tikzstyle{ap}  = [bp   , fill= yellow!20, text width= 14em, minimum height=19em, rounded corners]
\tikzstyle{q2g} = [ap   , text width= 22em, minimum height=5em, rounded corners]

\tikzstyle{obj}   = [bp   ,rounded corners, minimum height=1.5em]
\tikzstyle{det}   = [vis  ,rounded corners, text width=3em, minimum height=3em]

\def\blockdist{2.1}
\vspace{0.5ex}
\setlist{nolistsep}
\begin{enumerate}[leftmargin=*]
\setlength\itemsep{0.4mm}
\item \textbf{Question mapping into a graph representation -} First, a representation of the question as a directed graph is generated, where nodes represent objects and edges represent relations between objects. Graph components include objects classes, properties and relations. 
The node representation includes all the object visual requirements needed to answer the question, which is a combination of the following \scalebox{0.95}[1.0]{(see examples in the supplement, section 1)}:

\vspace{0.5ex}
 \begin{itemize}[leftmargin=*]
\setlength\itemsep{0.4mm}
    \item Object class $c$ (\eg 'horse').
    \item Object property $p$ (\eg 'red').
    \item Queried object property $f$ (\eg 'color').
    \item Queried set property $g$ (\eg 'number').
    \item Quantifiers (\eg 'all', 'two').  
    \item Quantity relative to another node (\eg same).
    \item Node type: regular or SuperNode: OR of nodes (with optional additional requirements).
\end{itemize}




\vspace{0.5ex}
\item \textbf{Answering procedure - } In this stage, a recursive procedure finds valid assignments of the graph in the image. The number of required assignments for each node is determined by its quantifiers. The procedure follows the graph, invoking relevant sub-procedures and integrates the information to provide the answer. Importantly, it depends only on the abstract structure of the question graph, where the particular object classes, properties and relations are parameters, used to apply the corresponding visual estimators (\eg which property to extract). The invoked sub-procedures are selected from a pool of the following \textit{basic procedures}, which are simple visual procedures used to compose the full answering procedure: 
   \begin{itemize}[leftmargin=*]
\setlength\itemsep{0.4mm}
\vspace{0.4ex}
    \item Detect object of a certain class $c$.
    \item Check the existence of object property $p$.
    \item Return an object property of type $f$.
    \item Return an object's set property of type $g$.
    \item Check the existence of relation $r$ between two objects. 
    \end{itemize}
\end{enumerate}
\vspace{0.4ex}

Our construction of a question graph and using its abstract structure to guide the answering procedure leads to our ability to handle novel domains by adding visual estimators but using the same answering procedure.
In our method we only train the question-to-graph mappers and the required visual estimators. Unlike QA training, we use independent trainings, which may utilize existing methods and be developed separately. This also simplifies domain extension (\eg automatic modification is simpler for question-graph examples than for question-image-answer examples). 
\subsection{Question to Graph Mapping}
\label{sec:mapping}
\vspace{-1ex}

Understanding natural language questions and parsing them to a logical form is a hard problem, still under study \cite{jia2016data,andreas2013semantic,wang2015building,berant2013freebase,ringgaard2017sling}. Retrieving question's structure by language parsers was previously performed in visual question answering \cite{andreas2016neural}, utilizing the Stanford Parser \cite{klein2003accurate}. 

We handled the question-to-graph task as a translation problem from natural language questions into a graph representation, training an LSTM based sequence to sequence models \cite{seq2seqNIPS2014}. The graph was serialized (using DFS traversal) and represented as a sequence of strings (including special tokens for graph fields), so the model task is to translate the question sequence into the graph sequence (see examples in Section 1 of the supplement). All our models use the architecture of Google's Neural Machine Translation model \cite{wu2016google}, and are trained using tensorflow implementation \cite{luong17}. 
A simple post-processing fixes invalid graphs. The description below starts with a question-to-graph model trained for CLEVR data, and then elaborates on the generation of extended models, trained for extended scopes of questions.



\vspace{-2ex}
\subsubsection{Question-to-Graph for CLEVR Data}
\vspace{-1ex}

Our basic question-to-graph model is for CLEVR questions and categories (3 objects, 12 properties, 4 property types, 4 relations). The graph annotations are based on the CLEVR answering programs \cite{johnson2016clevr}, corresponding to the dataset's questions. The programs can be described as trees, where nodes are functions performing visual evaluations for object classes, properties and relations. These programs can be transferred to our graph representation, providing annotations for our mappers training. Note that concepts may be mapped to their synonyms (\eg 'ball' to 'sphere').

\vspace{-2ex}
\subsubsection{Extended Question-to-Graph Domain}
\label{sec:mappingExt}
\vspace{-1ex}

CLEVR questions are limited, both in the used categories and in question types (\eg without quantifiers). To handle questions beyond the CLEVR scope, we trained question-to-graph mappers using modified sets of questions (randomization was shown to enable domain extension \cite{tobin2017domain}). There were two types of modifications: increasing the vocabulary of visual elements (object classes, properties and relations) and adding questions of new types. The vocabulary was expanded by replacing CLEVR visual elements with ones from a larger collection. This operation does not add question examples to the set, but uses the existing examples with replaced visual elements. Note that as this stage deals with question mapping and not question answering, the questions, which are generated automatically, do not have to be meaningful (\eg "What is the age of the water?") as long as they have a proper mapping, preserving the role of each visual element. 
To guarantee graph-question correspondence a preprocessing is performed where for each concept, all its synonyms are modified to one form. In addition, for each question all appearances of a particular visual element are replaced with the same term. We used three replacement 'modes', each generating a modified dataset by selecting from a corresponding set (real world categories from existing datasets): 
\begin{inparaenum}[i)]
\item \textbf{Minimal:} Most categories are from COCO \cite{lin2014microsoft} and VRD \cite{lu2016visual} (100 objects, 32 properties, 7 property types, 82 relations).
\item \textbf{Extended:} 'Minimal' + additional categories, sampled from 'VG' (230 objects, 200 properties, 53 property types, 160 relations).
\item \textbf{VG:} The categories of the Visual Genome dataset \cite{krishna2017visual} (65,178 objects, 53,498 properties, 53 property types, 47,448 relations, sampled according to prevalence in the dataset). The categories include many inaccuracies, such as mixed categories (\eg 'fat fluffy clouds') and irrelevant concepts (\eg objects: 'there are white'), which adds inconsistency to the mapping.
\end{inparaenum}

The second type of question modification increased the variability of questions. We created enhanced question sets where additional examples were added to the sets generated by each replacement mode (including 'None'). These examples include questions where '\textit{same \textless$p$\textgreater}' is replaced with '\textit{different \textless$p$\textgreater}' (where \textless$p$\textgreater \hspace{0.2mm} is a property), questions with added quantifiers ('all' and numbers) and elemental questions (with and without quantifiers). The elemental questions were defined as existence and count questions for: class, class and property, class and 2 properties, 2 objects and a relation, as well as queries for objects class (in a relation) and property types (including various WH questions).

The words vocabulary we used for training all sets was the same: ~56,000 words, composed by the union of the English vocabulary from IWSLT'15 \cite{Luong-Manning:iwslt15} together with all the used object classes, properties and relations. Both the question and the graph representations were based on the same vocabulary, with additional tokens in the graph vocabulary to mark graph nodes and fields (\eg \textit{\textless NewNode\textgreater}, \textit{\textless p\textgreater}).


Different mappers were trained for all the modified sets above. An example of a graph, mapped using the 'Extended-Enhanced' model, as well as the corresponding original question is given in Figure \ref{fig:ExtEnGraph}. Note that the modified question, although meaningless, has the same structure as the original question and is mapped to the same graph, except for the replaced visual elements and added quantifiers.  
This means that the same answering procedure will be carried out, fulfilling our intent to apply the same procedure to similar structured questions.


\tikzstyle{cir5} = [draw, circle,fill=yellow!20, text centered, text width=4em, inner sep=0.5ex, scale=0.8]
\tikzstyle{line} = [draw, very thick, color=black!50, -latex']

\begin{figure} [h!]
\vspace{-2ex}
\begin{center}
\resizebox{\columnwidth}{!}{
\renewcommand{\arraystretch}{0.8}
\setlength\tabcolsep{1.5pt} 
\begin{tabular}{cc}
\begin{tikzpicture}[scale=3, node distance = 0.3cm, auto]
    \node [cir5, align = center, inner sep=0cm, scale=0.92] at ( 0.5, 0) (node4)   {$\boldsymbol{c}$:\hspace{0.3mm}object \\
                                                                      $\hspace{-1.2mm}\boldsymbol{p}$:\hspace{0.3mm}tiny,\hspace{0.3mm}red,\\
                                                                         metallic 
                                                                      };
    \node [cir5, align = center, inner sep=0cm] at (-0.5, 0) (node3) { $\boldsymbol{c}$:\hspace{0.3mm}sphere \\
                                                                       $\boldsymbol{p}$:\hspace{0.3mm}cyan } ; 

    \node [cir5, above of = node3, align = center, inner sep=0cm] at (0, 0.2) (node2) {$\boldsymbol{c}$: object \\
                                                                                       $\boldsymbol{f}$: size};
    \node at (-0.3, 0.21) {\scriptsize{right}};
    \node at (0.28, 0.21) {\scriptsize{left}};

    \path [line] (node4) -- (node2);
    \path [line] (node3) -- (node2);
\end{tikzpicture}
 &
\begin{tikzpicture}[scale=3, node distance = 0.3cm, auto]
    \node [cir5, align = left, inner sep=0cm, text width=4.8em, scale=0.78] at ( 0.5, 0) (node4) {\hspace{2mm}$\boldsymbol{c}$:\hspace{0.3mm}object \\
                                                                    \hspace{-1.6mm}\small{$\boldsymbol{p}$:}\hspace{0.3mm}\small{full,}\hspace{0.3mm}\small{tied-up,}\\
                                                                     \hspace{4mm}\hspace{0.5mm}tiled\\
                                                                    \vspace{-0.5mm}\hspace{4mm}$\boldsymbol{n}$:\hspace{0.3mm}16};
    \node [cir5, align = center, inner sep=0cm, text width=4.5em, scale=0.84] at (-0.5, 0) (node3)    {  $\boldsymbol{c}$:\hspace{0.3mm}girl \\
                                                                         $\hspace{-0.5mm}\boldsymbol{p}$:\hspace{0.3mm}\small{light\_blue}\\
                                                                                  $\boldsymbol{q}$:\hspace{0.3mm}all};

    \node [cir5, above of = node3, align = center, inner sep=0cm] at (0, 0.2) (node2) {$\boldsymbol{c}$:\hspace{0.3mm}object \\
                                                                                       $\boldsymbol{f}$:\hspace{0.3mm}fabric};
    \node at (-0.5, 0.26) {\scriptsize{walking\_towards}};
    \node at (0.35, 0.26) {\scriptsize{next\_to}};

    \path [line] (node4) -- (node2);
    \path [line] (node3) -- (node2);
\end{tikzpicture}
\\
\multicolumn{1}{l}{\scriptsize{\textbf{Q}: What is the size of the object that is both }}   & \multicolumn{1}{l}{\scriptsize{\textbf{Q}: What is the fabric of the object that is both }}  \\
\multicolumn{1}{l}{\scriptsize{\Aindent    right of the cyan sphere and left of the tiny }} & \multicolumn{1}{l}{\scriptsize{\Aindent    walking towards all the light blue girls and }}   \\
\multicolumn{1}{l}{\scriptsize{\Aindent    red metallic object?}}                           & \multicolumn{1}{l}{\scriptsize{\Aindent    next to the sixteen full tied-up tiled objects?}}
\end{tabular}
}
\caption[ExtEn graph]{Left: A CLEVR question and a corresponding graph. Right: A modified question and a corresponding graph, mapped using Extended-Enhanced model. The accuracy of the modified representation is confirmed, as it matches the original accurate graph (with modified graph concepts).} 
\label{fig:ExtEnGraph}
\end{center}
\vspace{-2ex}
\end{figure}

\vspace{-3ex}
\subsection{Answering Procedure}
\label{sec:procedure}
\vspace{-1ex}

In this stage a recursive procedure seeks valid assignments (see Section \ref{sec:overview}) between the question graph and the image. The question graph, the image and the mask R-CNN \cite{he2017maskrcnn} produced for the image provide the input to the procedure that recursively processes each node (see Figure \ref{fig:scheme}). For each node, basic procedures (see Section \ref{sec:overview}) are invoked sequentially, according to the node's requirements and activate visual estimators according to the particular visual elements. The number of required valid assignments is set by the node's quantifier (a single assignment, a specific number, or all) or by the need of all objects for evaluating the entire object set (\eg counting, number comparisons). The next processed nodes are according to a DFS traversal. Each basic procedure provides an answer, used to produce the final answer, reporting unsupported categories and providing elaborations, based on intermediate results. For more details and examples see Section 2 of the supplement. 

\vspace{-2ex}
\subsubsection{CLEVR Visual Estimators}
\label{sec:clevr_dets}
\vspace{-1ex}

In order to find a valid assignment of a question graph in the image, and provide the answer, corresponding visual estimators need to be trained. 
Object locations are not explicitly provided for CLEVR data, however they can be automatically recovered using the provided scene annotations. This process provided approximated contour annotations for CLEVR objects (see Figure \ref{fig:clevr_det}), which were used for training. Mask R-CNN \cite{he2017maskrcnn} was used for instance segmentation. For property classifiers, simple CNN models (3 convolutional layers and 3 fully connected layers) were trained to classify color and material; size was estimated according to object's bottom coordinates and its largest edge. Relations are classified according to objects' locations. 



\vspace{-2ex}
\subsubsection{Real World Visual Estimators}
\label{sec:real_dets}
\vspace{-1ex}

Handling questions in the real-world domain beyond CLEVR objects was performed by utilizing existing visual estimators. For instance segmentation we use a pre-trained mask R-CNN \cite{he2017maskrcnn} for the 80 classes of COCO dataset \cite{lin2014microsoft}. Any other visual estimator may be incorporated to enhance answering capability. In our experiments (Section \ref{sec:res_ext} and Figure \ref{fig:clevr_new}) we use color map estimation \cite{van2007learning}, age and gender classification \cite{LH:CVPRw15:age} (utilizing face detection \cite{Mathias2014Eccv}) and depth estimation \cite{Depth2015Liu} (utilized for estimating spatial relations).

\section{Experiments}
\label{sec:empirical}
\vspace{-1ex}

The experiments tested the abilities of the UnCoRd system, to first, provide accurate results for the CLEVR dataset and second, to handle extended questions and real-world domains. Our analysis included the two answering stages: creating a correct graph representation of the question, and answering the questions. Adam optimizer was used for question-to-graph and visual estimators training with a learning rate of $10^{-4}$ ($10^{-3}$ for the 'Extended-Enhanced' model), selected according to the corresponding validation set results. Each model training was using one NVIDIA Tesla V100 GPU. All reported results are for a single evaluation. For each model, the same version was used in all experiments.
Unless stated, system was configured to provide short answers (concise and without elaborations); markings on images in the figures correspond to intermediate results. 
Code will be available at \scalebox{0.74}[1.0]{\url{https://github.com/benyv/uncord}}

\vspace{-0.1ex}
\subsection{CLEVR Experiments}
\vspace{-0.7ex}

We trained a question-to-graph model for \scalebox{.98}[1.0]{CLEVR} ('None'-'Basic', as denoted in Section \ref{sec:map_res}), which generated 100\% perfect graphs on its validation set. The visual estimators, described in Section \ref{sec:clevr_dets} were also trained and provided the results given in Table \ref{table:clevrEst}. \scalebox{.98}[1.0]{CLEVR} relations were estimated by simple rules using the objects' coordinates. 




\vspace{-2ex}
\begin{table}[h!]
\setlength\tabcolsep{0.1pt} 
\begin{minipage}{0.64\columnwidth}
\centering
\begin{tabular}{cc}
    \subfigure{
    \includegraphics[width=2.5cm]{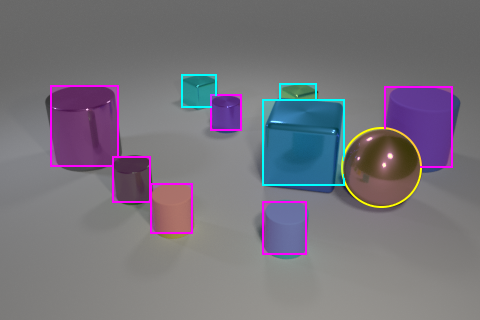} 
    } &
    \subfigure{
    \includegraphics[width=2.5cm]{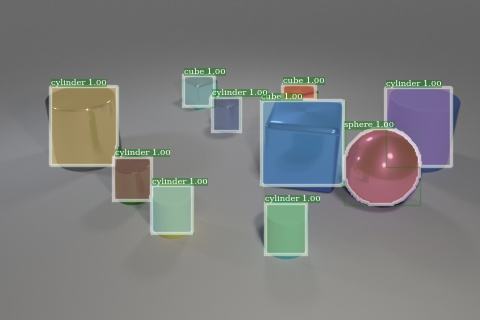} 
    }
\end{tabular}
\captionof{figure}{Instance segmentation example for CLEVR data. Left: GT (approximated from scene data), Right: results.}
\label{fig:clevr_det}
\end{minipage}\hfill
\begin{minipage}{0.335\columnwidth}
\centering
\resizebox{\columnwidth}{!}{
   \subfigure{
    \begin{tabular}{  l | c | c }
    \toprule
    \textbf{Estimator}                     &   \textbf{AP$^{\textbf{IoU=.50}}$}  &  \textbf{Acc.} \\ 
    \midrule
    \midrule
    \textbf{Ins. seg.}      &        99.0        &                    \\ 
    \textbf{Color}          &                    &        99.98       \\ 
    \textbf{Material}       &                    &        99.97       \\ 
    \textbf{Size}           &                    &         100        \\ 
   \bottomrule
   \end{tabular}
   }
   }
\caption[CLEVR estimators results]{CLEVR estimators results on CLEVR validation set}
\label{table:clevrEst}
\end{minipage}

\vspace{-1ex}
\end{table}

We tested the answering performance of the UnCoRd system on the CLEVR test set. The results, including for other state-of-the-art methods (all use answers labels for training) are given in Table \ref{table:clevrRes}.

\begin{table}[h!]
\vspace{-1ex}
\centering
\begin{minipage}{0.48\textwidth}
\begin{centering}
    \resizebox{\columnwidth}{!}{%
    \begin{tabular}{  l  c  c  c  c  c | c | c} 
    \toprule
    \multirow{2}{*}{\textbf{Method}} & \multirow{2}{*}{\scalebox{0.9}[1.0]{\textbf{Exist}}} & \multirow{2}{*}{\scalebox{0.9}[1.0]{\textbf{Count}}}  & \scalebox{0.9}[1.0]{\textbf{Comp.}} & \scalebox{0.9}[1.0]{\textbf{Query}}     & \scalebox{0.9}[1.0]{\textbf{Comp.}}   & \scalebox{0.9}[1.0]{{\textbf{Overall}}} & \scalebox{0.9}[1.0]{{\textbf{Overall}}}\\  
    &                                &                                 & \scalebox{0.9}[1.0]{\textbf{Num.}} & \scalebox{0.9}[1.0]{\textbf{Att.}} & \scalebox{0.9}[1.0]{\textbf{Att.}} &  \scalebox{0.9}[1.0]{\textbf{test set}} & \scalebox{0.9}[1.0]{\textbf{val. set}} \\
    \midrule
    \midrule
    \scalebox{0.9}[1.0]{\textbf{\scalebox{0.9}[1.0]{IEP-}strong} \cite{johnson2017inferring}}        & 97.1 & 92.7 & 98.7 & 98.1 & 98.9 & 96.9 & \\
    \scalebox{0.9}[1.0]{\textbf{FiLM} \cite{perez2018film}}                     & 99.3 & 94.3 & 93.4 & 99.3 & 99.3 & 97.6 & \\
    \scalebox{0.9}[1.0]{\textbf{\scalebox{0.9}[1.0]{DDR}prog} \cite{suarez2018ddrprog}}              & 98.8 & 96.5 & 98.4 & 99.1 & 99.0 & 98.3 & \\
    \scalebox{0.9}[1.0]{\textbf{MAC} \cite{hudson2018compositional}}            & 99.5 & 97.1 & 99.1 & 99.5 & 99.5 & 98.9 & \\
    \scalebox{0.9}[1.0]{\textbf{TbD} \cite{mascharka2018transparency}}          & 99.2 & 97.6 & 99.4 & 99.5 & 99.6 & 99.1 & \\
    \scalebox{0.9}[1.0]{\textbf{HAN} \cite{malinowski2018learning}}             & 99.6 & 97.2 & 96.9 & 99.6 & 99.6 & 98.8 & \\
    \scalebox{0.85}[1.0]{\textbf{NS-VQA} \cite{nsvqa}\footnote{Reported for val. set, hence not compared to test set results}} & 99.9 & 99.7 & 99.9 & 99.8 & 99.8 & - & \textbf{99.8} \\
    \midrule
    \scalebox{0.9}[1.0]{\textbf{UnCoRd\textsubscript{None-B}}}                                        & \textbf{99.89} & \textbf{99.54} & \textbf{99.91} & \textbf{99.74} & \textbf{99.80} & \textbf{99.74} & \textbf{99.8} \\ 
 \bottomrule
   \end{tabular}
   }
\caption[clevr results]{CLEVR QA accuracy for state-of-the-art methods}\vspace{-1ex}
\label{table:clevrRes}
\end{centering}
\end{minipage}
\vspace{-2ex}
\end{table}


As can be seen, our model achieves state-of-the-art results without training for the visual question answering task and not using any answers GT, as other methods. In addition UnCoRd can elaborate and explain answers and failures using intermediate results, and extend the handled domain with no need of images and related QA examples, as demonstrated in Section \ref{sec:domainExt} and Figure \ref{fig:clevr_ex}. On a sample of 10,000 validation set examples, all mistakes were due to wrong visual estimators' predictions, mainly miss detection of a highly occluded object. Hence, accurate annotation of object coordinates (as performed in NS-VQA \cite{nsvqa}) may even further reduce the small number of errors. Note that NS-VQA requires full scene analysis, which is not scalable for domain extension with a large number of objects and relations. It also uses images with question-answer pairs to train the programs, coupling the method to the specific trained question answering domain. 

\subsection{Out of Domain Experiments}
\label{sec:domainExt}
\vspace{-1ex}

Next, we test UnCoRd beyond the scope of CLEVR data. We trained question-to-graph models on the modified and enhanced CLEVR data and used corresponding visual estimators. We examined whether domain extension is possible while maintaining a good performance on the original data.

\vspace{-2ex}
\subsubsection{Question to Graph}
\label{sec:map_res}
\vspace{-1ex}

For evaluating question representation, we trained and tested (see Section \ref{sec:mappingExt}) 8 question-to-graph models that include all replacement modes (None, Minimal, Extended, VG), each trained in two forms: Basic (B), \ie no added question examples ($\smallsim$700K examples) and Enhanced (E), \ie with additional examples ($\smallsim$1.4M examples). 

In Table \ref{table:seq2seqResFull}, we report the results of each trained model on the validation sets of all 8 models, which provides information on generalization across the different sets. Note that as the "None" extension, unlike the data of other models, includes mapping from concepts to their synonyms (see Section \ref{sec:mappingExt}), prediction for "None" data by the "Minimal", "Extended" and "VG" models include a preprocessing stage transforming each concept synonyms to a single form. 


\begin{table}[h!]
\vspace{-1ex}
\centering
\begin{centering}
    \resizebox{\columnwidth}{!}{%
    \begin{tabular}{  l  c | c  c | c  c | c  c | c  c } 
    \toprule
    \multicolumn{2}{c|}{\multirow{2}{*}{\diagbox{\textbf{Train}}{\textbf{Test}}}}  & \multicolumn{2}{c|}{\textbf{None}} &  \multicolumn{2}{c|}{\textbf{Minimal}} & \multicolumn{2}{c|}{\textbf{Extended}} & \multicolumn{2}{c}{\textbf{VG}}\\  
    \multicolumn{2}{ l|}{} & \textbf{B} & \textbf{E} & \textbf{B} & \textbf{E} & \textbf{B} & \textbf{E} &  \textbf{B} & \textbf{E}\\ 
    \midrule
    \midrule
    \multirow{2}{*}{\textbf{None}}     &  \textbf{B} &  100 & 49.5 &  0.5 &  0.2 &  0.1 &  0.0 &  0.1 &  0.1  \\ 
                                       &  \textbf{E} & 99.7 & 99.8 &  0.5 &  0.4 &  0.1 &  0.1 &  0.1 &  0.1 \\ 
    \midrule
    \multirow{2}{*}{\textbf{Minimal}}  &  \textbf{B} & 99.8 & 48.9 & 98.4 & 50.0 &  0.5 &  0.3 &  1.2 &  0.6  \\ 
                                       &  \textbf{E} & 99.0 & 98.6 & 98.0 & 97.7 &  0.5 &  1.0 &  1.1 &  1.1  \\ 
    \midrule
    \multirow{2}{*}{\textbf{Extended}} &  \textbf{B} & 99.1 & 48.6 & 98.2 & 49.9 & 96.2 & 49.1 & 18.1 &  9.4  \\ 
                                       &  \textbf{E} & 99.1 & 98.7 & 97.9 & 97.5 & 95.7 & 95.8 & 19.3 & 20.0  \\ 
    \midrule
    \multirow{2}{*}{\textbf{VG}}       &  \textbf{B} & 87.5 & 44.8 & 65.7 & 34.6 & 84.1 & 45.3 & 76.9 & 41.9  \\ 
                                       &  \textbf{E} & 90.0 & 90.0 & 63.7 & 64.1 & 81.9 & 83.0 & 75.0 & 77.1  \\ 
   \bottomrule
   \end{tabular}
   }
\caption[Q2G results across data]{Accuracy of question-to-graph mapping for all data types} 

\label{table:seq2seqResFull}
\end{centering}
\vspace{-1ex}
\end{table}

Results demonstrate that models perform well on data with lower variability than their training data. The high performance of the 'Extended' models on their corresponding data illustrates that substantial extensions are possible in question-to-graph mapping without requiring any new training images. 
VG models' lower accuracy is expected due to the unsuitable elements in its data (see Section \ref{sec:mappingExt}). Additional tests are required to check possible advantages of VG models for different domains. We report such a test next.



\vspace{-2ex}
\subsubsection{VQA Representation}
\label{sec:vqa_rep}
\vspace{-1ex}

In this experiment, representation capabilities are tested for a different dataset. Since normally, annotations corresponding to our graph representation are not provided, we sampled 100 questions of the VQA \cite{VQA} validation set and manually examined the results for the eight question-to-graph models (see Section \ref{sec:map_res}).


The results in Table \ref{table:vqaRes} express the large gaps in the abilities of models to represent new domains. Models trained specifically on CLEVR do not generalize at all to the untrained domain. As the models are trained on more diverse data, results improve substantially, peaking clearly for VG-Enhanced model by a large margin from other models. This is also evident in the example given in Figure \ref{fig:vqa-graphs} where adequacy of the graph increases in a similar manner. This result is interesting as using this model provides high accuracy for CLEVR as well (see Table \ref{table:clevrResUncord}). The fact that substantial performance gain is achieved for a data domain that was not used in training (the VQA dataset domain), while preserving good results on the original data (CLEVR), demonstrates the potential of the approach to provide a general answering system for visual questions. Further investigation is required for means to enrich question description examples and produce further significant improvements.

\tikzstyle{cir1} = [draw, circle,fill=yellow!20, text centered, text width=5.2em, inner sep=0.1mm, scale=0.8]

\begin{table} [h!]
\centering
    \begin{tabular}{ c  c | c  c | c  c | c  c } 
    \toprule
    \multicolumn{2}{c|}{\textbf{None}} &  \multicolumn{2}{c|}{\textbf{Minimal}} & \multicolumn{2}{c|}{\textbf{Extended}} & \multicolumn{2}{c}{\textbf{VG}}\\  
    \textbf{B} & \textbf{E} & \textbf{B} & \textbf{E} & \textbf{B} & \textbf{E} &  \textbf{B} & \textbf{E}\\ 
    \midrule
    \midrule
       1       &    0       &     12     &      12    &    22      &      22       &      34        &      50      \\ 
    \bottomrule
    \end{tabular}
\caption[Q2G vqa]{Accuracy of graph representation for VQA \cite{VQA} sample, given for the different UnCoRd mappers. As expected, training on more diverse data allows better generalization across domains.}
\label{table:vqaRes}
\end{table}
\vspace{-2ex}

\begin{figure} [h!]
\centering
\vspace{-0.5ex}
\resizebox{0.95\columnwidth}{!}{
\begin{tabular}{ccc}
\multicolumn{3}{c}{\vspace{-1.5ex}\textbf{Q}: \textit{What kind of ground is beneath the young baseball player?}}\\
\vspace{-1.2ex}
\subfigure{
\begin{tikzpicture}[scale=1.5, node distance = 2.1cm, auto]
    \node [cir1, align = left] (node1) { $ \boldsymbol{c}$:  {\color{red} cylinder} };

    \node [cir1, above of = node1, align = left] at (0, 0.2) (node2) {$\quad \boldsymbol{c}$: {\color{red} cube }\\
                                                                      $      \boldsymbol{f}$: {\color{red} material} };
    \node at (-0.57, 0.65) { {\color{red} 'same size'}};
    \path [line] (node1) -- (node2);
\end{tikzpicture}
}\hspace{0em}
\subfigure{
\begin{tikzpicture}[scale=1.5, node distance = 2.1cm, auto]
    \node [cir1, align = left] (node1) { $   \boldsymbol{c}$: {\color{red} baseball } \\
                                         $ \; \boldsymbol{p}$: {\color{blue} young }};
    \node [cir1, above of = node1, align = left] at (0, 0.2) (node2) {$\;\; \boldsymbol{c}$: {\color{red} what} };
    \node at (-0.47, 0.68) {{\color{blue} 'beneath'}};
    \path [line] (node1) -- (node2);
\end{tikzpicture}
}\hspace{0em}
\subfigure{
\begin{tikzpicture}[scale=1.5, node distance = 2.1cm, auto]
    \node [cir1, align = left] (node1) { $ \boldsymbol{c}$: {\color{blue} baseball \\ \quad\; player} \\
                                         $ \; \boldsymbol{p}$: {\color{blue} young} };
    \node [cir1, above of = node1, align = left] at (0, 0.2) (node2) {$ \; \boldsymbol{c}$: {\color{blue}ground} \\
                                                                      $ \;\;  \boldsymbol{f}$: {\color{blue}kind} };
    \node at (-0.47, 0.68) {{\color{blue} 'beneath'}};
    \path [line] (node1) -- (node2);
\end{tikzpicture}
} \\
\multicolumn{3}{l}{\hspace{2.8em} \small{\textbf{None-Basic}} \hspace{1.7em}  \small{\textbf{Min-Enhanced}} \hspace{1.3em}  \small{\textbf{VG-Enhanced}}}  

%

\end{tabular}
}
\caption[VQA graph examples]{Generated graphs for a free form question (from the VQA \cite{VQA} dataset). Blue text: accurate concepts, red: inaccurate.} 
\label{fig:vqa-graphs}
\vspace{-2ex}
\end{figure}
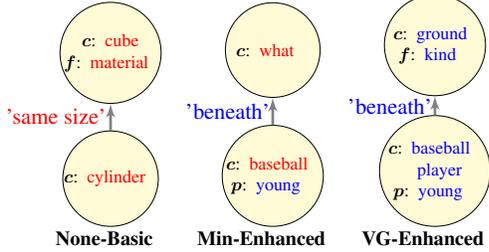

\vspace{-2ex}
\subsubsection{Maintaining Performance on CLEVR Questions}
\label{sec:clevr_res_uncord}
\vspace{-1ex}

We evaluated the performance change for the CLEVR test set, as the training data variability of the question-to-graph models increases. The results are given in Table \ref{table:clevrResUncord}.

\begin{table}[h!]
\vspace{-1ex}
\begin{centering}
    \resizebox{\columnwidth}{!}{%
    \begin{tabular}{  l  c  c  c  c  c  c | c } 
    \toprule
    \multicolumn{2}{c}{\multirow{2}{*}{\textbf{Mapper}}} & \multirow{2}{*}{\textbf{Exist}} & \multirow{2}{*}{\textbf{Count}}  & \textbf{Comp.} & \textbf{Query}     & \textbf{Comp.}   & \multirow{2}{*}{\textbf{Overall}}\\  
                                &                        &                                 &                                  & \textbf{Num.} & \textbf{Att.} & \textbf{Att.} & \\
    \midrule
    \midrule

    \multirow{2}{*}{\textbf{None}} &  \textbf{B}  &  \textbf{99.89} & \textbf{99.54} & \textbf{99.91} & \textbf{99.74} & \textbf{99.80} & \textbf{99.74}  \\ 
                                          &  \textbf{E}  &  \textbf{99.89} & \textbf{99.54} & \textbf{99.91} & \textbf{99.74} & \textbf{99.80} & \textbf{99.74}  \\ 
    \multirow{2}{*}{\textbf{Min}}  &  \textbf{B}  & 99.81 & 99.36 & 99.87 & 99.73 & \textbf{99.80} & 99.68 \\ 
                                          &  \textbf{E}  & 99.69 & 99.21 & 99.47 & 99.46 & 99.59 & 99.46 \\ 
    \multirow{2}{*}{\textbf{Ext}}  &  \textbf{B}  & 96.82 & 89.34 & 78.64 & 99.40 & 99.41 & 94.80 \\ 
                                          &  \textbf{E}  & 99.78 & 99.33 & 98.36 & 99.65 & 99.76 & 99.49 \\ 
    \multirow{2}{*}{\textbf{VG}}   &  \textbf{B}  & 96.82 & 89.34 & 78.64 & 99.44 & 99.41 & 94.81 \\ 
                                          &  \textbf{E}  & 98.03 & 97.39 & 96.88 & 97.62 & 97.22 & 97.49 \\ 
 \bottomrule
   \end{tabular}
   }
\caption[clevr UnCoRd results]{Accuracy of CLEVR dataset question answering by UnCoRd using the different question-to-graph mappers}
\label{table:clevrResUncord}
\end{centering}
\vspace{-1ex}
\end{table}

\begin{figure} [h!]
\vspace{-1ex}
\begin{centering}
\resizebox{\columnwidth}{!}{%
\begin{tabular}{llll}
\subfigure{
  \includegraphics[totalheight=0.16\textheight]{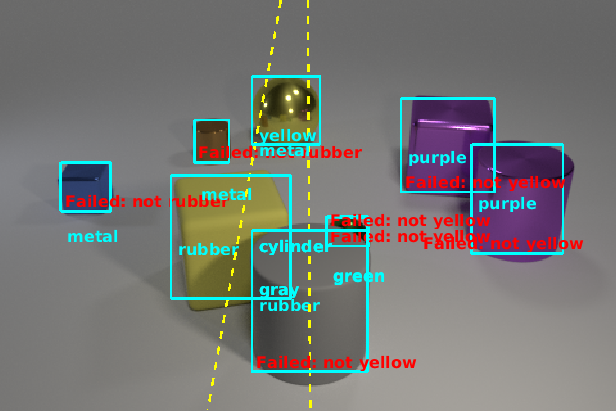} 
} &
\subfigure{
  \includegraphics[totalheight=0.16\textheight]{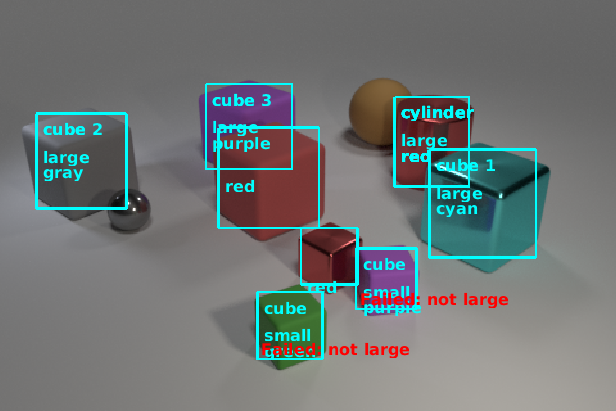} 
}  \\
(a) \textbf{Q}: There is a yellow thing to the      & (b) \textbf{Q}: How many big cubes are of a       \\
\aindent \Aindent right of the rubber thing on the  & \aindent \Aindent \textit{different} color than the large  \\
\aindent \Aindent left side of the gray rubber      & \aindent \Aindent cylinder?                       \\
\aindent \Aindent  cylinder; what is its material?  & \aindent \textbf{A}: 3                            \\
\aindent \textbf{A}:  metal                         &                                                   \\

\subfigure{
  \includegraphics[totalheight=0.16\textheight]{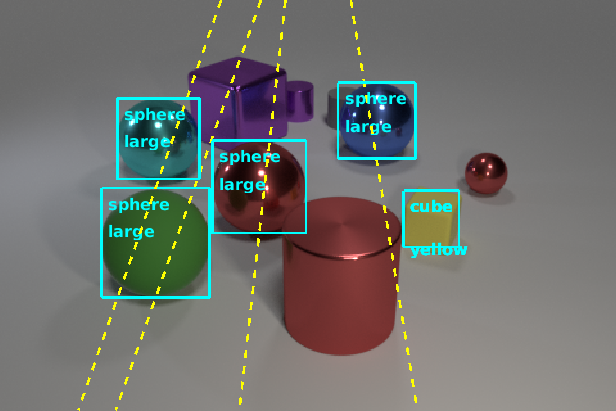} 
} &
\subfigure{
  \includegraphics[totalheight=0.16\textheight]{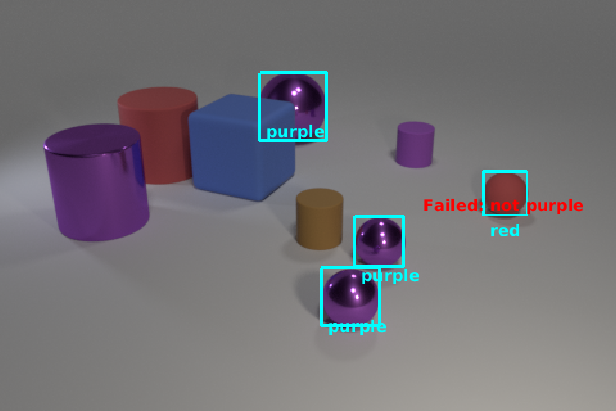} 
}  \\
(c) \textbf{Q}: What color is the cube to the     & (d) \textbf{Q}: \textit{Are all the spheres purple?} \\
\aindent \Aindent right of the \textit{four} big spheres?  & \aindent \textbf{A}: no \small{[full: There are not enough purple }   \\
\aindent \textbf{A}: yellow                       & \aindent \Aindent \quad \small{spheres (failed due to a red sphere)]} \\[6pt]

\end{tabular}
}
\vspace{-1.5ex}
\caption[CLEVR QA examples]{Examples for answering different question types on CLEVR images: (a) taken from CLEVR, (b) includes 'different color' relation, (c) uses a quantifier, and (d) a simple property existence (+ 'all' quantifier) question.}
\label{fig:clevr_ex}
\end{centering}
\vspace{-2.7ex}
\end{figure}

It is evident that even models that were trained on a much larger vocabulary and question types than the original CLEVR data still perform well, mostly with only minor accuracy reduction. This demonstrates that with more variable training we can handle more complex questions, while maintaining good results on the simpler domains. 
Examples on CLEVR images for both CLEVR questions and others are shown in Figure \ref{fig:clevr_ex} (using 'None-Enhanced' mapper). 

\vspace{-1ex}
\subsubsection{CLEVR Humans}
\label{sec:clevr_hum}
\vspace{-1ex}

An example of using the CLEVR images with different questions is the CLEVR-Humans \cite{johnson2017inferring} (7145 questions in test set), where people were asked to provide challenging questions for CLEVR images. The questions vary in phrasing and in the required prior knowledge. 

\begin{table} [h!]
\vspace{-2ex}
\begin{minipage}{0.5\columnwidth}
\centering
    \resizebox{\columnwidth}{!}{%
    \begin{tabular}{  l  c | c c } 
    \toprule
    \multicolumn{2}{ l|} {\textbf{Method}}            & \textbf{No FT}   & \textbf{FT}\\  
    \midrule
    \midrule
    \multicolumn{2}{ l|} {\textbf{IEP-18k}}                     &  54.0        &   66.6   \\
    \multicolumn{2}{ l|} {\textbf{FiLM}}                        &  56.6        &   75.9   \\
    \multicolumn{2}{ l|} {\textbf{MAC}}                         &  57.4        &   \textbf{81.5}   \\
    \multicolumn{2}{ l|} {\textbf{NS-VQA}}                      &   -          &   67.0   \\
    \midrule

    \multirow{2}{*}{\textbf{\small{UnCoRd-None}}} &  \hspace{-0.3cm} \textbf{B}  &  60.46         &  \\ 
                                          &  \hspace{-0.3cm} \textbf{E}  & \textbf{60.59} &  \\ 
    \multirow{2}{*}{\textbf{\small{UnCoRd-Min}}}  &  \hspace{-0.3cm} \textbf{B}  & 48.24          &   \\ 
                                          &  \hspace{-0.3cm} \textbf{E}  & 52.23          &   \\ 
    \multirow{2}{*}{\textbf{\small{UnCoRd-Ext}}}  &  \hspace{-0.3cm} \textbf{B}  & 43.97          &   \\ 
                                          &  \hspace{-0.3cm} \textbf{E}  & 52.83          &   \\ 
    \multirow{2}{*}{\textbf{\small{UnCoRd-VG}}}   &  \hspace{-0.3cm} \textbf{B}  & 43.47          &   \\ 
                                          &  \hspace{-0.3cm} \textbf{E}  & 48.71          &   \\ 
   \bottomrule
   \end{tabular}
   }
\caption[clevr humans results]{Question answering accuracy of CLEVR-Humans test set for state-of-the-art methods, with and without finetuning (FT).}
\label{table:humansRes}
\end{minipage}\hfill
\begin{minipage}{0.5\columnwidth}
\centering
\vspace{-0.5ex}
\resizebox{\columnwidth}{!}{%
\begin{tabular}{l}
\subfigure{
  \includegraphics[totalheight=0.16\textheight]{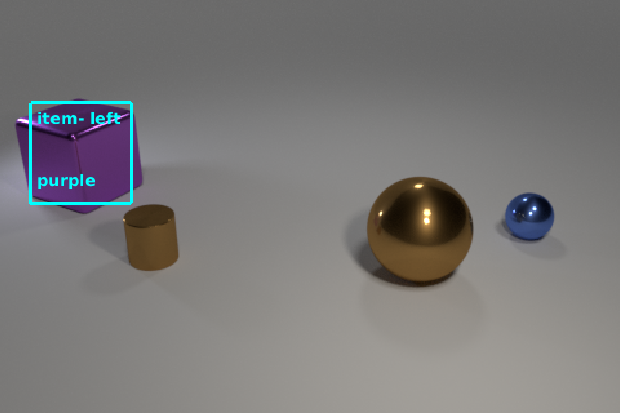} 
} \vspace{-0.5ex} \\
\textbf{Q}: What color is the item to the far left?     \\
\Aindent          {\color{blue}\textit{(GT: purple)}}   \\
\textbf{None-E \ A}: brown, \textbf{VG-E A}: purple       \\
\scalebox{.78}[1.0]{\textbf{IEP-Str A} (No FT): blue, \textbf{IEP-Hum A} (FT): purple}  \vspace{-1ex}\\
\subfigure{
  \includegraphics[totalheight=0.16\textheight]{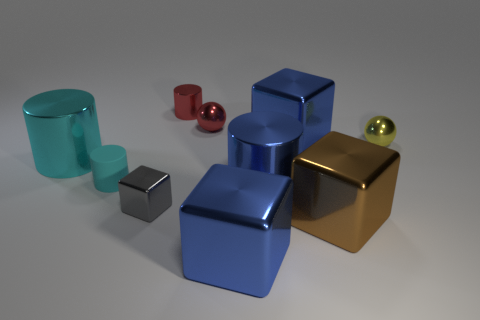} 
}  \vspace{-0.5ex} \\
\textbf{Q}: How many of these things could be           \\
\Aindent    stacked on top of each other? {\color{blue}\textit{(GT: 8)}} \\
\scalebox{.8}[1.0]{\textbf{None-E A}: 1, \textbf{VG-E A}: Unknown class: each other} \\
\scalebox{.8}[1.0]{\textbf{IEP-Str A} (FT): 0, \textbf{IEP-Hum A} (FT): 2}  \\[6pt]
\end{tabular}
}
\vspace{-1.5ex}
\captionof{figure}[CLEVR-Humans examples]{Examples for CLEVR-Humans questions} 
\label{fig:clevr_hum}
\end{minipage}
\vspace{-1.5ex}
\end{table}

Results, given in Table \ref{table:humansRes}, demonstrate that for models without finetuning, our 'None-Enhanced' model provides state-of-the-art results (without any answer examples). For all models, questions with phrasing not included in training are prone to errors, including 'hallucinations' of concepts. Note that CLEVR-Humans answers are the same answers as in CLEVR (by instructions to workers), hence models biased towards CLEVR (the "None" models) have a better success chances. Models with a rich vocabulary may capture the question graph more accurately, but that may include concepts with no corresponding visual estimators, resulting with answers such as: "Unknown relation 'in between'". Adding such visual estimators will improve performance. Since accuracy calculation does not reward for such limitation indications, just "guessing" the answer would increase the computed accuracy, especially as success chances rise with a simple question categorization (\eg 50\% for yes/no and size questions). However, indicating limitations gives a better sense of the system's level of understanding the question, and can lead to corrective actions. Such answers can be promoted in QA systems, by reducing "score" for wrong answers, or giving partial scores to answers identifying a missing component.

Examples of CLEVR-Humans questions are given in Figure \ref{fig:clevr_hum}. 
It is evident that the more general model (VG-Enhanced) can perform on out of scope questions (top) and report limitations (bottom).

\vspace{-1.8ex}
\subsubsection{Extensibility to Real-World Images}
\label{sec:res_ext}
\vspace{-1ex}

The UnCoRd system can be naturally extended to novel domains by a simple plug-in of visual estimators. This is illustrated in Figure \ref{fig:clevr_new} for using new properties/relations and for an entirely different domain of real-world images. An experiment that adds questions with a novel property is presented in Section 3 of the supplement. We next describe an experiment for real-world images, where we use real world visual estimators (see Section \ref{sec:real_dets}) and our most general trained mapper (VG-Enhanced). We compare our model to Pythia \cite{pythia18arxiv}, which has top performance on the VQA v2 dataset \cite{balanced_vqa_v2}. The experiment includes two parts:
\vspace{0.5ex}
\begin{enumerate}[leftmargin=*]
\setlength\itemsep{0.4mm}
  \item \scalebox{0.96}[1.0]{'Non VQA\_v2'} questions: 100 questions outside Pythia's training domain (VQA v2), with unambiguous answers, on 50 COCO images (two similar questions per image with different answers). We freely generated questions to include one or more of the following categories: 
\vspace{0.5ex}
  \begin{itemize}[leftmargin=*]
\setlength\itemsep{0.4mm}
    \item A combination of properties and relations requirements linked by logical connectives ('and', 'or').
    \item Property comparison (\eg 'same color'). 
    \item Quantifiers (\eg 'all', 'five').
    \item Quantity comparison (\eg 'fewer', 'more'). 
    \item A chain of over two objects connected by relations.
  \end{itemize}
  \item 'VQA\_v2' questions: 100 questions sampled from VQA v2 dataset \cite{balanced_vqa_v2} with terms that have visual estimators in UnCoRd and unambiguous answers (annotated by us). 
\end{enumerate}

In addition to the estimators mentioned in Section \ref{sec:real_dets}, ConceptNet \cite{Speer2013} is used by UnCoRd to query for optional classes when superordinate groups are used (\eg 'animals'). More details are in Section 4 of the supplement.  

\begin{figure} [h!]
\vspace{-2ex}
\begin{centering}
\resizebox{\columnwidth}{!}{%
\begin{tabular}{ll}
\multicolumn{1}{c}{\subfigure{
  \includegraphics[totalheight=0.14\textheight]{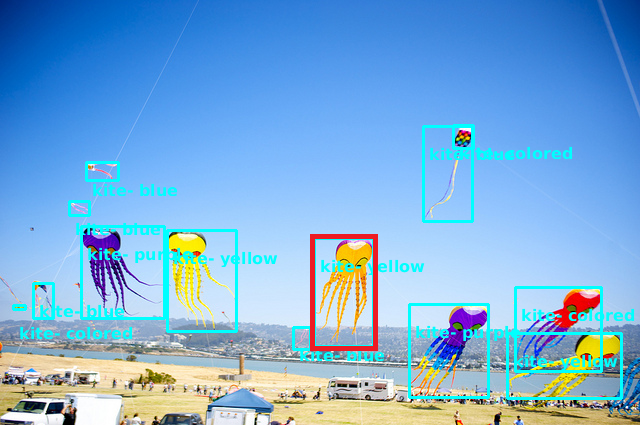}
}} &
\multicolumn{1}{c}{\subfigure{
  \includegraphics[totalheight=0.14\textheight]{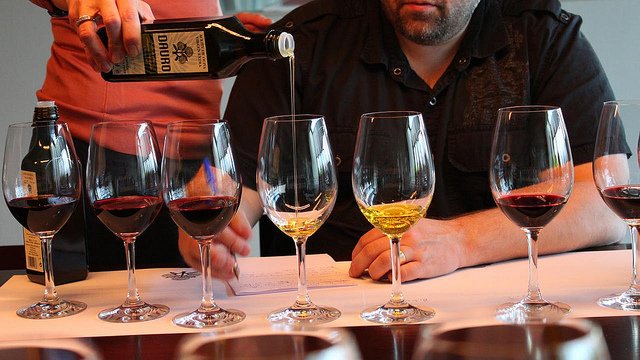}
}}  \\
\textbf{Q}: What color is the kite closest to the  & 
\textbf{Q}: There is a bottle that is left of a bottle; \\
\Aindent    yellow kite that is left of the orange kite? & 
\Aindent    how many wine glasses are right of it?      \\
\scalebox{.88}[1.0]{\textbf{UnCoRd A}: {\color{red} no valid orange kite}, \textbf{Pythia A}: {\color{red} blue}} & \textbf{UnCoRd A}: {\color{red} no bottle}, \textbf{Pythia A}: {\color{red} 5}  
                                                         \\
\textbf{Q}: What color is the kite closest to the  & 
\textbf{Q}: There is a bottle that is right of a bottle;\\
\Aindent    yellow kite that is right of the       & 
\Aindent    how many wine glasses are left of it?       \\
\Aindent    orange kite?                           &  
\textbf{UnCoRd A}: {\color{red} no bottle}, \textbf{Pythia A}: {\color{red} 5}   \\
\scalebox{.88}[1.0]{\textbf{UnCoRd A}: {\color{red} no valid orange kite}, \textbf{Pythia A}: {\color{red} blue}}  & 
                                                        \\
\hline
\noalign{\vskip 0.5ex}
\textbf{Failure source}: wrong color estimation \cite{van2007learning} & 
\textbf{Failure source}:  failed object detection       \\
(below: failed object in the red box above) & 
(below: mask R-CNN results \cite{he2017maskrcnn})       \\
\multicolumn{1}{c}{\begin{tabular}{@{}l@{}}Object \\ pixels \\ \\ \end{tabular}
 \subfigure{
  \includegraphics[totalheight=0.14\textheight]{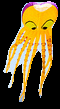} 
}
\subfigure{
  \includegraphics[totalheight=0.14\textheight]{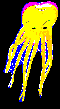} 
}
\begin{tabular}{@{}l@{}}\scalebox{.88}[1.0]{Predicted} \\ pixels' \\ colors \\ (yellow \\ instead \\ \scalebox{.88}[1.0]{of orange)} \\ \\ \\ \\ \\ \\  \end{tabular}
} 
& 
\multicolumn{1}{c}{\subfigure{
  \includegraphics[totalheight=0.14\textheight]{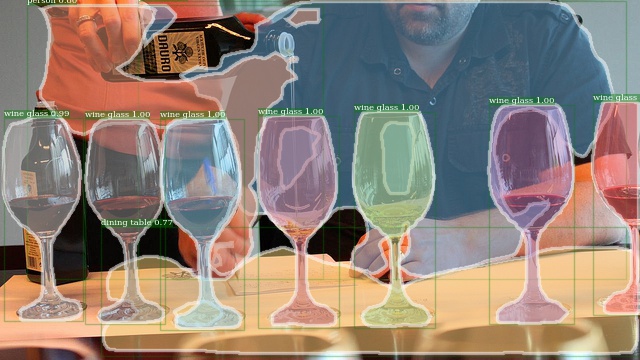}
}}  \\
\end{tabular}
}
\vspace{-8.5ex}
\caption[Examples of question answering failures for novel questions on natural images]{Examples of UnCoRd failures in answering questions outside the VQA v2 domain on COCO images.} 
\label{fig:coco_bad}
\end{centering}
\vspace{-3ex}
\end{figure}

The non VQA\_v2 results, given in Table \ref{table:coco-comp}, demonstrate the substantial advantage of UnCoRd for these types of questions. All UnCoRd's failures are due to wrong results of the invoked visual estimators. Note the substantial performance difference in Pythia between yes/no and WH questions, unlike the moderate difference in UnCoRd. We found that Pythia recognizes the yes/no group (\ie answers 'yes'/'no'), but its accuracy (56\%) is close to chance level (50\%). Examples of successful UnCoRd answers to the non VQA\_v2 questions are provided in Figure \ref{fig:coco_good}, while failure examples, including failure sources, are shown in Figure \ref{fig:coco_bad}. Pythia's answers are given as well.


\begin{table}[h!]
\vspace{-1.4ex}
\begin{centering}
    \begin{tabular}{  l c c | c } 
    \toprule
    \textbf{Method}                      & \textbf{Yes/No}  &  \textbf{WH}  & \textbf{Overall} \\  
    \midrule
    \midrule
    \textbf{Pythia \cite{pythia18arxiv}} &       56.0       &     14.0      &       35.0       \\
    \midrule
    \textbf{UnCoRd-VG-E}                 &  \textbf{88.0}   & \textbf{64.0} &  \textbf{76.0}   \\ 
   \bottomrule
   \end{tabular}
\caption[Question answering results for novel questions on natural images]{Answering accuracy for 100 questions outside the VQA v2 domain (including quantifiers, comparisons, multiple relation chains and multiple relations and properties) on COCO images.}
\label{table:coco-comp}
\end{centering}
\vspace{-2.3ex}
\end{table}


\definecolor{green2}{rgb}{0.0, 0.5, 0.0}
\begin{figure} [h!]
\vspace{-1ex}
\begin{centering}
\resizebox{\columnwidth}{!}{%
\begin{tabular}{ll}
\multicolumn{1}{c}{\subfigure{
  \includegraphics[totalheight=0.18\textheight]{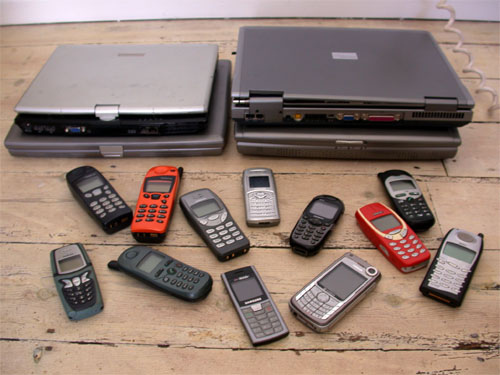}
}} &
\multicolumn{1}{c}{\subfigure{
  \includegraphics[totalheight=0.18\textheight]{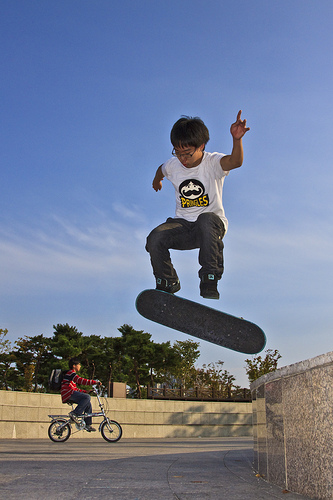}
}} \\
\textbf{Q}: How many cell phones are left of the     & \textbf{Q}: What object is supporting the person   \\
\Aindent    red cell phone that is closest to the    & \Aindent    that is left of the person above the   \\
\Aindent    right cell phone?                        & \Aindent    skateboard?                            \\
\textbf{UnCoRd A}: {\color{green2} 9}, \textbf{Pythia A}: {\color{red} 4} & \textbf{UnCoRd A}: {\color{green2} bicycle}, \textbf{Pythia A}: {\color{red} skateboard} \\
\textbf{Q}: How many cell phones are left of the     & \textbf{Q}: What thing is on an object that is left\\
\Aindent    right cell phone?                        & \Aindent    of the person above the skateboard? \\
\textbf{UnCoRd A}: {\color{green2} 11}, \textbf{Pythia A}: {\color{red} 5} & \textbf{UnCoRd A}: {\color{green2} person}, \textbf{Pythia A}: {\color{red} skateboard}  \\
\multicolumn{1}{c}{\subfigure{
  \includegraphics[totalheight=0.18\textheight]{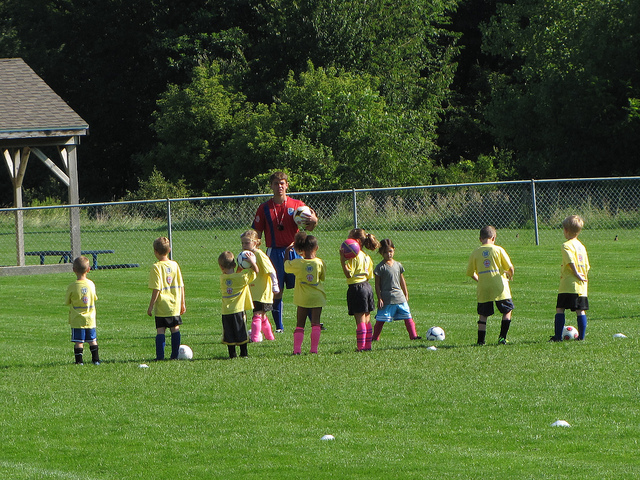}
}} &
\multicolumn{1}{c}{\subfigure{
  \includegraphics[totalheight=0.18\textheight]{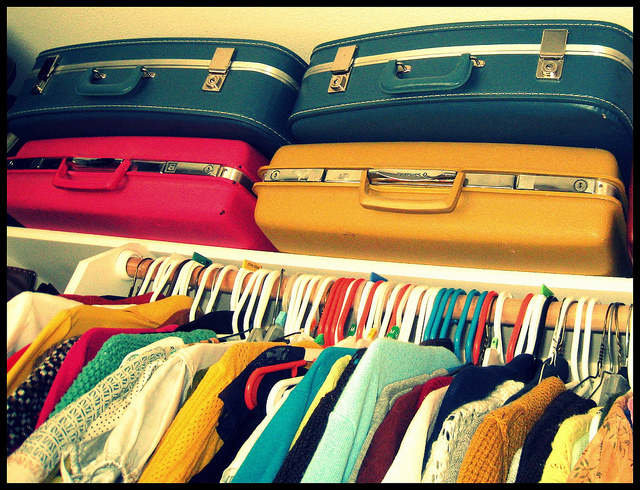}
}}  \\
\textbf{Q}:Is the number of people that are to the  & \textbf{Q}: What color is the suitcase that is both     \\
\Aindent   right of the left ball the same as the   & \Aindent    below a blue suitcase and left of a         \\
\Aindent   number of balls?                         & \Aindent    suitcase?                                   \\
\textbf{UnCoRd A}: {\color{green2} no}, \textbf{Pythia A}: {\color{green2} no} & \textbf{UnCoRd A}: {\color{green2} red}, \textbf{Pythia A}: {\color{red} blue} \\
\textbf{Q}: Is the number of people that are to the & \textbf{Q}: What color is the suitcase that is both     \\
\Aindent    right of the left ball greater than the & \Aindent    below a blue suitcase and right of a        \\
\Aindent    number of balls?                        & \Aindent    suitcase?                                   \\
\textbf{UnCoRd A}: {\color{green2} yes}, \textbf{Pythia A}: {\color{red} no} & \textbf{UnCoRd A}: {\color{green2} orange}, \textbf{Pythia A}: {\color{red} blue} \\
\end{tabular}
}
\caption[Examples of question answering successes for novel questions on natural images]{Examples of UnCoRd successes in answering questions outside the VQA v2 domain on COCO images.}
\label{fig:coco_good}
\end{centering}
\vspace{-1ex}
\end{figure}


\begin{figure} [h!]
\vspace{-2ex}
\begin{centering}
\resizebox{\columnwidth}{!}{%
\begin{tabular}{lll}
\multicolumn{1}{c}{\subfigure{
  \includegraphics[totalheight=0.18\textheight]{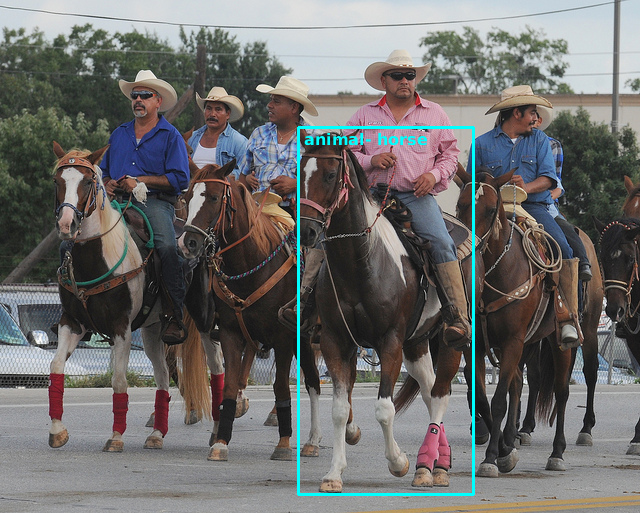}
}}  &
\multicolumn{1}{c}{\subfigure{
  \includegraphics[totalheight=0.18\textheight]{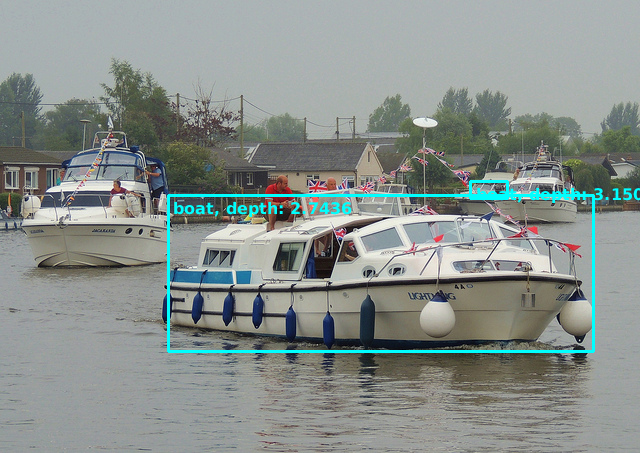}
}} \\
\textbf{Q}: What kind of animal is shown?                                                & \textbf{Q}: Is there a boat behind the car?        \\
\textbf{VG-E A}: {\color{green2} horse} \small{[\textbf{full:} The type of the animal: } & \textbf{VG-E A}: {\color{green2} no} \small{[\textbf{full:} There is no car} \\
\leavevmode\hphantom{\textbf{VG-E A}: horse} \small{horse, where horse is a }            & \leavevmode\hphantom{\textbf{VG-E A}: no \small{[\textbf{full:}}}\small{ The boat is behind a boat]}  \\
\leavevmode\hphantom{\textbf{VG-E A}: horse} \small{subclass of animal]}                 & \textbf{Pythia A}: {\color{red} yes}               \\
\textbf{Pythia A}: {\color{green2} horse}                                                &  \vspace{-0.5ex}
\\
\multicolumn{1}{c}{\subfigure{
  \includegraphics[totalheight=0.18\textheight]{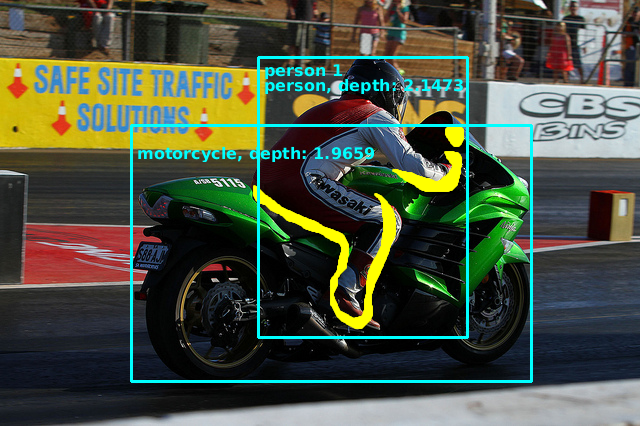}
}} &
\multicolumn{1}{c}{\subfigure{
  \includegraphics[totalheight=0.18\textheight]{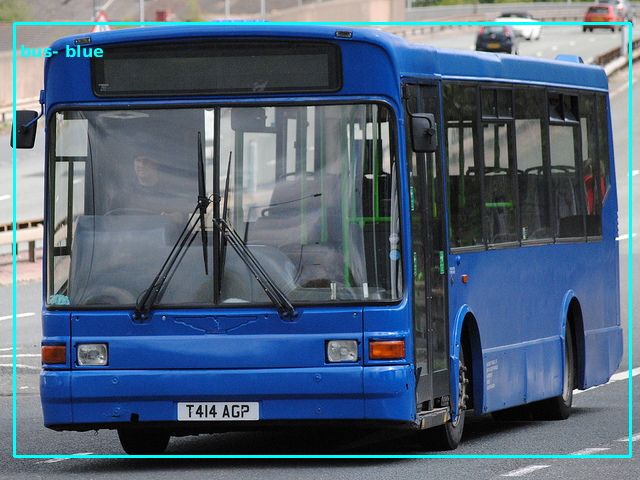}
}}  \\
\textbf{Q}:How many people are on the motorcycle?   & \textbf{Q}: Is there a yellow bus in the picture?       \\
\textbf{UnCoRd A}: {\color{green2} 1}               & \textbf{UnCoRd A}: {\color{green2} no} \small{[\textbf{full:} There are no yellow} \\
\textbf{Pythia A}: {\color{green2} 1}               & \leavevmode\hphantom{\textbf{UnCoRd A}: no} \small{buses (failed due to a blue bus)]} \\
                                                    & \textbf{Pythia A}: {\color{green2} no}                  \vspace{-0.5ex}
\\
\multicolumn{1}{c}{\subfigure{
  \includegraphics[totalheight=0.18\textheight]{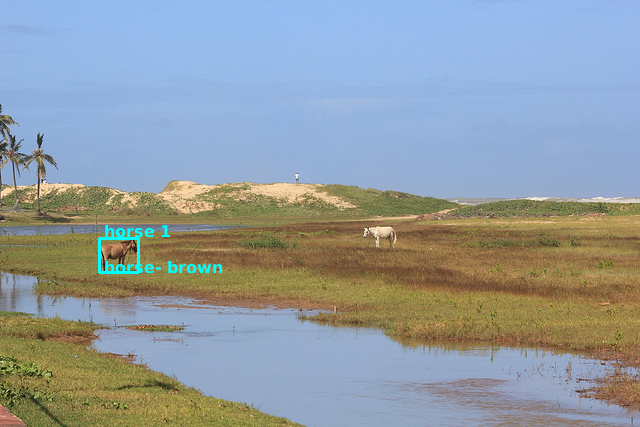}
}} &
\multicolumn{1}{c}{\subfigure{
  \includegraphics[totalheight=0.18\textheight]{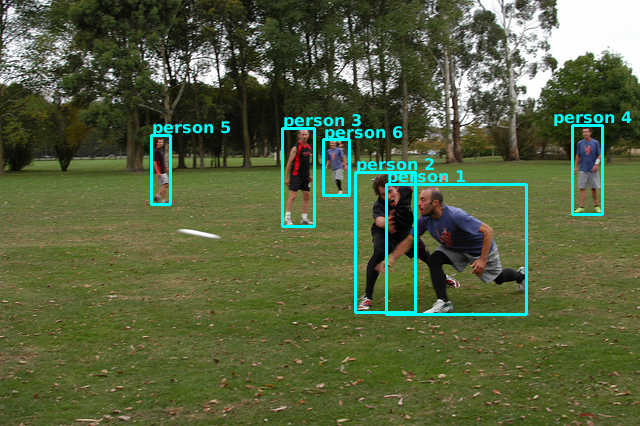}
}}  \\
\textbf{Q}: How many brown horses are there?       & \textbf{Q}: How many people are there?              \\ 
 \textbf{UnCoRd A}: {\color{green2} 1}, \textbf{Pythia A}: {\color{red} 2} & \textbf{UnCoRd A}: {\color{green2} 6}, \textbf{Pythia A}: {\color{red} 7} \\ 
\end{tabular}
}
\caption[Examples of question answering for natural images sampled from VQA v2 dataset]{Examples of UnCoRd answers to VQA v2 questions (including 'full' answers when they add information).} 
\label{fig:vqa2}
\end{centering}
\vspace{-2.9ex}
\end{figure}

Results for the 100 VQA\_v2 questions are given in Table \ref{table:coco-vqa2}. As can be seen, UnCoRd's results are better by a large margin, compared to Pythia \cite{pythia18arxiv} end-to-end model, even though questions were sampled from VQA v2, a dataset used for Pythia's training. As in the previous part, all UnCoRd's failures are only due to wrong results of the invoked visual estimators. Examples of UnCoRd's answers for the VQA\_v2 questions are given in Figure \ref{fig:vqa2}, including the corresponding answers of Pythia. 

\begin{table}[h!]
\vspace{-1ex}
\begin{centering}
    \begin{tabular}{  l c c | c } 
    \toprule
    \textbf{Method}                       & \textbf{Yes/No}  &  \textbf{WH}  & \textbf{Overall} \\  
    \midrule
    \midrule
    \textbf{Pythia \cite{pythia18arxiv}}  &       90.0       &      68.3     &       77.0       \\
    \midrule
    \textbf{UnCoRd-VG-E}                  &  \textbf{97.5}   & \textbf{88.3} &  \textbf{92.0}  \\ 
   \bottomrule
   \end{tabular}
\caption[VQA v2 question answering results]{Answering accuracy for 100 questions sampled from VQA v2 dataset (on terms with visual estimators in UnCoRd).}
\label{table:coco-vqa2}
\end{centering}
\vspace{-0.5ex}
\end{table}


The above experiments on real-world images show that when corresponding visual estimators are available, our method performs better than a leading end-to-end model, both for questions outside the training domain of the end-to-end model (where the advantage is substantial) and for questions from this domain. This is achieved without any question answering training.



\section{Conclusions and Future Directions}
\label{conclusion}
\vspace{-1ex}


We proposed a novel approach to answer visual questions by combining a language step, which maps the question into a graph representation, with a novel algorithm that maps the question graph into an answering procedure. Because the algorithm uses the abstract structure of this graph, it allows a transfer to entirely different domains. Training is performed for the language step to learn the graph representation, and for the visual step to train visual estimators. However, unlike existing schemes, our method does not use images and associated question-answer pairs for training. Our approach allows handling novel domains provided that corresponding visual estimators are available. The combination of the question graph and answering procedure also gives the method some capacity to explain its answers and suggest alternatives when question is not grounded in the image. 
Based on this approach, our answering system achieves near perfect results on a challenging dataset, without using any question-answer examples. We have demonstrated that question representation and answering capabilities can be extended outside the scope of the data used in training, preserving good results for the original domain.

Substantial work is required to obtain a system that will be able to perform well on entirely general images and questions. The main immediate bottleneck is obtaining question-to-graph mapping with general representation capabilities for a broad range of questions. 
Question graph representation may also be enhanced to support questions with more complex logic, as well as extending the scope of the supported visual categories (\eg global scene types). Any general VQA requires vast estimation capabilities, as any visual category can be queried. In UnCoRd they are modularly incremented and automatically integrated with existing questions.
Additional basic areas that current schemes, including ours, have only begun to address, are the use of external, non-visual knowledge in the answering process, and the composition of detailed, informative answers, integrating the language and visual aspects of VQA.

\vspace{-2.5ex}
\paragraph*{Acknowledgements:} This work was supported by EU Horizon 2020 Framework 785907 and ISF grant 320/16.

{\small
\bibliographystyle{ieee_fullname}
\bibliography{ref}
}

\end{document}